\documentclass[lettersize,journal]{IEEEtran}
\usepackage{amsmath,amsfonts}
\usepackage{algorithmic}
\usepackage[ruled]{algorithm2e}
\usepackage{array}
\usepackage[caption=false,font=normalsize,labelfont=sf,textfont=sf]{subfig}
\usepackage{textcomp}
\usepackage{stfloats}
\usepackage{url}
\usepackage{verbatim}
\usepackage{multirow}
\usepackage{graphicx}
\def\BibTeX{{\rm B\kern-.05em{\sc i\kern-.025em b}\kern-.08em
    T\kern-.1667em\lower.7ex\hbox{E}\kern-.125emX}}
\usepackage{balance}
\begin{document}
\title{Deep Intra-Image Contrastive Learning for Weakly Supervised One-Step Person Search}

\author{Jiabei~Wang,
        Yanwei~Pang,~\IEEEmembership{Senior~Member,~IEEE},
        Jiale~Cao,
        Hanqing~Sun,
        Zhuang~Shao,
        and~Xuelong~Li,~\IEEEmembership{Fellow,~IEEE}
\thanks{J. Wang, Y. Pang, J. Cao, and H. Sun are with the School of Electrical and Information Engineering, Tianjin University, Tianjin 300072, China (E-mail: \{jiabeiwang,pyw,connor,hqSun\}@tju.edu.cn).}
\thanks{Z. Shao is with Warwick Manufacturing Group, University of Warwick, Coventry, CV4 7AL, UK (E-mail: Zhuang.Shao@warwick.ac.uk).}
\thanks{X. Li is with the School of Computer Science and Center for OPTical IMagery Analysis and Learning (OPTIMAL), Northwestern Polytechnical University, Xi'an 710072, P.R. China (E-mail: li@nwpu.edu.cn).}}

\markboth{Journal of \LaTeX\ Class Files,~Vol.~18, No.~9, September~2020}%
{How to Use the IEEEtran \LaTeX \ Templates}

\maketitle

\begin{abstract}

Weakly supervised person search aims to perform joint pedestrian detection and re-identification (re-id) with only person bounding-box annotations. Recently, the idea of contrastive learning is initially applied to weakly supervised person search, where two common contrast strategies are memory-based contrast and intra-image contrast. We argue that current intra-image contrast is shallow, which suffers from spatial-level and occlusion-level variance. In this paper, we present a novel deep intra-image contrastive learning using a Siamese network. Two key modules are spatial-invariant contrast (SIC) and occlusion-invariant contrast (OIC). SIC performs many-to-one contrasts between two branches of Siamese network and dense prediction contrasts in one branch of Siamese network. With these many-to-one and dense contrasts, SIC tends to learn discriminative scale-invariant and location-invariant features to solve spatial-level variance. OIC enhances feature consistency with the masking strategy to learn occlusion-invariant features. Extensive experiments are performed on two  person search datasets CUHK-SYSU and PRW, respectively. Our method achieves a state-of-the-art performance among weakly supervised one-step person search approaches. We hope that our simple intra-image contrastive learning can provide more paradigms on weakly supervised person search. The source code is available at \url{https://github.com/jiabeiwangTJU/DICL}.
\end{abstract}


\section{Introduction}
\IEEEPARstart{P}{erson} search is aimed at searching a given query person in a set of gallery images, which is regarded as a joint task of pedestrian detection  and  re-identification (re-id) \cite{Xiao-cuhk-2017-CVPR,Cao_PedSurvey_TPAMI_2022,Ye_ReID_TPAMI_2022}. Currently, most existing person search methods \cite{Chen-MGTS-2020-TIP,Zhao-CANR-2022-TCSVT,Cao-PSTR-2022-CVPR} are fully-supervised, which are developed based on person bounding-box annotations and identity annotations. Compared to the bounding-box annotations of pedestrian detection, the identity annotations of person re-id are substantially expensive and time consuming. On the one hand, it is difficult and expensive to obtain a large number of identity annotations across different scenes. There are nearly 72.7\% and 17.4\% person bounding boxes without identity annotations in CUHK-SYSU and PRW  datasets \cite{Han-RSiamNets-2021-ICCV}, respectively. On the other hand, it is time consuming to accurately annotate the persons with same identity across different scenes, especially when the person has a large variance in appearance, scale, occlusion, \textit{etc}. Therefore, it is necessary to explore person search with only bounding-box annotations, termed as weakly supervised person search.

\begin{figure}[t!]
\centering
\includegraphics[width=0.99\columnwidth]{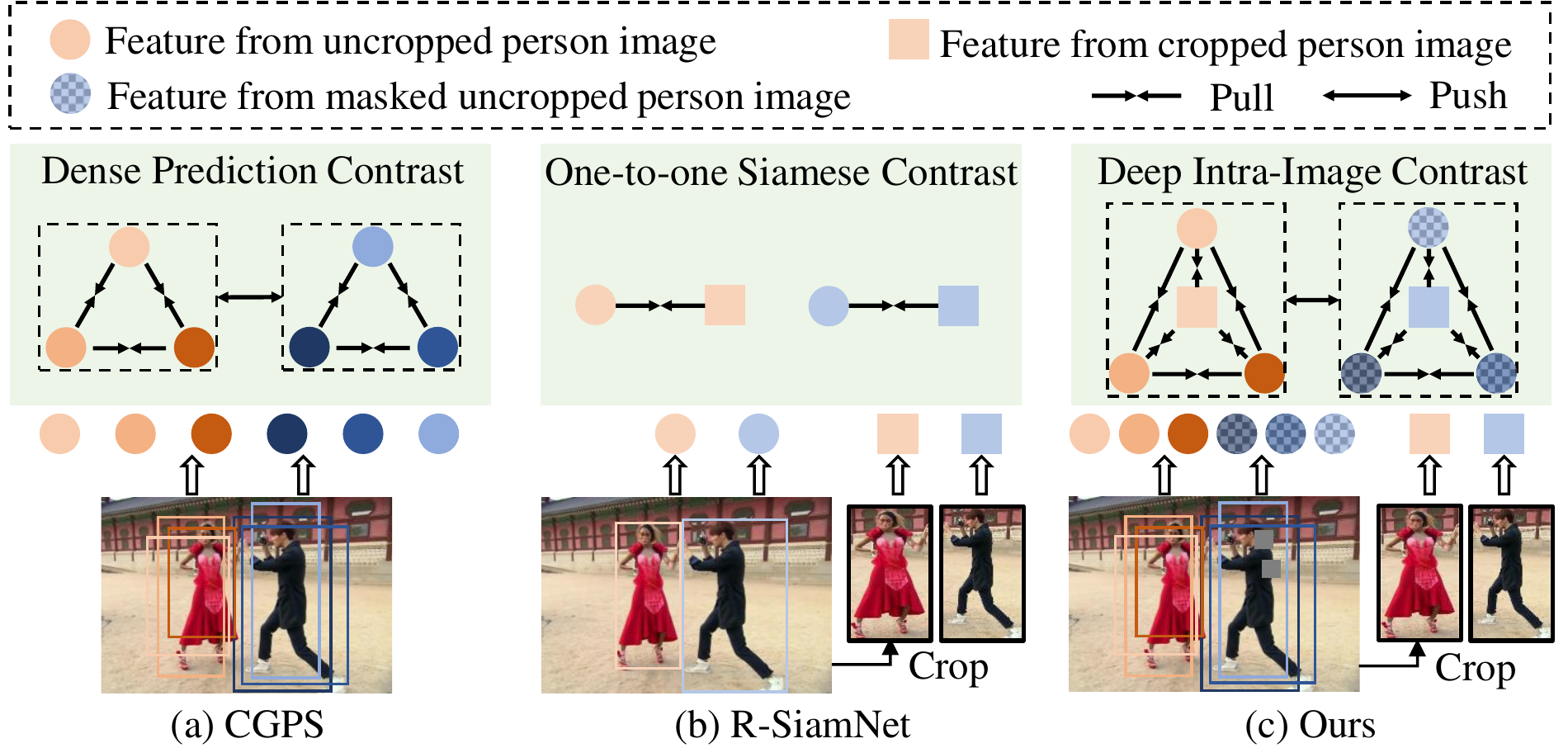}
\caption{Intra-image contrast strategies of different weakly supervised one-step  methods. CGPS (a) performs dense prediction contrasts using a single network with the input of an entire image. R-SiamNet (b) performs a one-to-one ground-truth contrast using Siamese network with the inputs of an entire image and cropped persons. Compared to these shallow intra-image contrast strategies, our method (c) exploits deep intra-image contrastive learning. We consider many-to-one Siamese contrasts between two branches of Siamese network and dense prediction contrasts in  one branch of Siamese network. In addition, we perform occlusion-invariant contrast by randomly masking a portion of person in one branch of Siamese network.}
\label{fig:compare}
\end{figure}

Recently, several contrastive learning based one-step methods have been proposed to perform weakly supervised person search. Two contrastive learning strategies are used in these methods, including memory-based contrast and intra-image contrast\footnote{Intra-image contrast can also mean intra-batch contrast during training. For simplicity, we use intra-image contrast here.}. Memory-based contrast builds a cluster-level re-id memory bank over all training images, and employs it to supervise the re-id feature learning in current image. Intra-image contrast aims to exploit the useful contrast information within image for re-id feature learning. We aim to improve the intra-image part. Fig. \ref{fig:compare} shows intra-image contrasts used in two typical weakly-supervised one-step methods CGPS \cite{Yan-CGPS-2022-AAAI} and R-SiamNet \cite{Han-RSiamNets-2021-ICCV}. As shown in Fig. \ref{fig:compare}(a), CGPS explores dense prediction contrasts using a single network. It enforces the re-id features belonging to one person be close, and pushes the features belonging to different persons apart. In Fig.  \ref{fig:compare}(b), R-SiamNet adopts a Siamese network for one-to-one Siamese contrast. It extracts two re-id features of a ground-truth in two branches and enforces these two re-id features with or without context be consistent. These methods achieve an initial success on weakly supervised one-step person search. However, it still lags far behind fully supervised one-step person search approaches \cite{Yan-AnchorFree-2021-CVPR,Chen-HOIM-2020-AAAI}.

We argue that one of key reasons is the insufficient  information mining with shallow intra-image contrast, which cannot effectively deal with \textit{spatial-level} and \textit{occlusion-level} variance of person instance. The detected person in different images could have a large variance in scale or an inaccurate bounding-box localization (spatial-level variance). In addition, the detected person  is sometimes occluded by surrounding objects (occlusion-level variance). As a result, it becomes difficult to match the re-id features of a same person across images due to these spatial-level and occlusion-level variance.

To address this issue caused by spatial-level and occlusion-level variance, we present a novel deep intra-image contrastive learning based method (DICL) that fully exploits instance-level information of each person in an image (Fig. \ref{fig:compare}(c)). Our DICL is built on a Siamese network with two novel contrast modules:  spatial-invariant contrast (SIC)  and occlusion-invariant contrast (OIC). In the Siamese network, 
a search branch extracts the re-id features from an entire image, while an instance branch extracts the re-id features from cropped ground-truths. Our SIC not only enforces many-to-one feature consistency between two branches of Siamese network, but also conducts dense prediction contrasts in the search branch. With many-to-one and dense contrasts, our SIC can learn scale-invariant and location-invariant features to deal with spatial-level variance. 
To deal with occlusion-level variance, our OIC performs the masking strategy in the search branch and enhances re-id feature consistency between two branches. Experimental results on two classical person search datasets CUHK-SYSU \cite{Xiao-cuhk-2017-CVPR} and PRW \cite{Zheng-prw-2017-CVPR} show the effectiveness and superiority of proposed DICL.
Our  contributions are summarized as follows:

\begin{itemize}
    \item We propose a novel weakly supervised one-step person search framework via deep intra-image contrastive learning (DICL). Without the identity annotations, DICL fully mines the intra-image information to deal with spatial-level and occlusion-level variance.
    \item We design two intra-image contrast modules: spatial-invariant  contrast (SIC) and occlusion-invariant contrast (OIC). SIC performs  many-to-one Siamese contrast and dense prediction contrast to learn scale-invariant and location-invariant features. OIC learns occlusion-invariant features with  masking strategy in search branch.  
    \item Our method achieves favorable results on CUHK-SYSU and PRW, which outperforms previous weakly supervised one-step approaches. On CUHK-SYSU test set, it achieves the mAP of 87.4\% and top-1 accuracy of 88.8\%. On PRW test set, it achieves the mAP of 35.5\% and top-1 accuracy of 80.9\%. 
\end{itemize}

\section{Related Work}
In this section, we first introduce person search, including fully/weakly supervised person search. After that, we give a brief review on unsupervised person re-identification (re-id)  and contrastive learning.

\subsection{Person Search}
Person search is a joint task of pedestrian detection and person re-id, which  plays an important role
in video surveillance \cite{Ke-JSE-2022-TCSVT,Yang-BUFF-2022-TCSVT,Gao-TPS-2022-TIP,Chen-CMKA-2021-TIP}. Existing person search methods can be divided into two main categories: two-step methods and one-step methods. Two-step methods \cite{Chen-MGTS-2018-ECCV,Wang-TCTS-2020-CVPR} detect pedestrians first and then identify them in the cropped person images, which perform two sub-tasks separately. Zheng \textit{et al.} \cite{Zheng-prw-2017-CVPR} explored different detectors and re-id methods for person search.  Chen \textit{et al.} \cite{Chen-MGTS-2018-ECCV,Chen-MGTS-2020-TIP} argued that person foreground region plays more important role for re-id and proposed a mask-guided two-stream CNN model  to emphasize the foreground information. Yao \textit{et al.} \cite{Yao-JPOR-2021-TIP} proposed to consider the objectness and repulsion information for similarity computation. Wang \textit{et al.} \cite{Wang-TCTS-2020-CVPR,Wang-BTCL-2021-TMM} employed an identity-guided query detector to generate query-like proposals and designed a detection results adapted re-ID model for improved re-id.

One-step methods \cite{Yan-AnchorFree-2021-CVPR,Chen-HOIM-2020-AAAI} perform pedestrian detection and person re-id in a unified end-to-end framework. Compared to two-step methods, one-step methods are simple but challenging, which needs to deal the diverse goals of pedestrian detection and person re-id. Most existing one-step methods are built on object detection framework Faster R-CNN \cite{Ren-FasterRCNN-2017-PAMI,Lin_FPN_CVPR_2017}. Xiao \textit{et al.} \cite{Xiao-cuhk-2017-CVPR} made an initial attempt by adding a fully-connected layer after R-CNN head network and generating the re-id features for online instance matching. Chen \textit{et al.} \cite{Chen-NormAware-2020-CVPR} proposed to decompose detection and re-id features in the polar coordinate system. Li \textit{et al.} \cite{Li-seqnet-2021-AAAI} employed a cascaded head-networks for detection and re-id, where the detection is first performed and the re-id is second performed. Han \textit{et al.} \cite{Han-DMRNet-2021-AAAI,Han-DMRNet-2022-TPAMI} proposed to decouple  detection and re-id by using RoI features only for  re-id. Dong \textit{et al.} \cite{Dong-BINet-2020-CVPR} introduced a
bi-directional interaction network taking both entire  images and the cropped instance images as inputs during training, using accurate human appearance information from person patches to discriminate identities. Han \textit{et al.} \cite{Han-AGWF-2021-ICCV}  perform
detection, re-identification, and part classification together for improved person search. In addition, some 
 other methods focus on improving online instance matching loss, such as HOIM \cite{Chen-HOIM-2020-AAAI} and OIM++ \cite{Lee-OIMNet++-2022-ECCV}. Instead of adopting anchor-based detector Faster R-CNN, Yan \textit{et al.} \cite{Yan-AnchorFree-2021-CVPR} proposed to adopt anchor-free detector FCOS \cite{Tian-FCOS-2019-ICCV} for anchor-free person search, which avoids the hand-crafted anchor design. 
Two recent works PSTR \cite{Cao-PSTR-2022-CVPR} and COAT \cite{Yu-COAT-2022-CVPR} employ the transformer-based network for person search.



\textbf{Weakly supervised person search:} Recently, several methods focus on weakly supervised person search. 
For example, two-step method CGUA \cite{Jia-CGUA-2022-ArXiv} proposes a context-guided cluster algorithm and unpaired-assisted memory to take advantage of plenty of unpaired persons. One-step method CGPS \cite{Yan-CGPS-2022-AAAI} exploits detection context for dense prediction contrast using a single network. R-SiamNet \cite{Han-RSiamNets-2021-ICCV} employs sparse ground-truths for intra-image contrast  with a Siamese network. We argue that these intra-image contrasts in one-step methods are shallow, which does not fully exploit intra-image information. 

\subsection{Unsupervised Person Re-ID}

Unsupervised person re-id learns discriminative features to identify different persons in unlabeled cropped images. Most deep learning-based  methods focus on the unsupervised domain adaption (UDA) task, which is with/without ID labels in source domain and without ID labels in target domain. The methods based on pseudo labels use clustering algorithm to classify instance features, such as k-means, DBSCAN \cite{Ester-DBSCAN-1996-KDD}, and Finch \cite{Sarfraz-finch-2019-CVPR}. For example, BUC \cite{Lin-BottomUp-2019-AAAI} treats each individual image as an identity  and applies a bottom-up clustering to the feature embedding reducing the number of classes. SPCL \cite{Ge-SPCL-2020-NeurIPS} proposes a self-paced contrastive learning strategy with a novel clustering reliability criterion to generate more reliable clusters. Some methods utilize cross-domain transfer learning \cite{Deng-SPGAN-2018-CVPR, Liu-ATNet-2019-CVPR}. DAPS \cite{Li-DAPS-2022-ECCV} presents a unsupervised domain adaptive person search paradigm.

\subsection{Contrastive Learning}
Contrastive learning learns feature representation from positive or negative data pairs. SimCLR \cite{Chen-SimCLR-2020-ICML} introduces a learnable non-linear transformation head to train two augmentation paths with large batch sizes and long training steps. BYOL \cite{Grill-BYOL-2020-NeurIPS} proposes an online and target two interactive networks slowly updating with only positive pairs. SimSiam \cite{Chen-SimSiam-2021-CVPR} represents a simple Siamese network without negative sample pairs, large batches or momentum encoders, using a stop-gradient operation to prevent collapsing. ContrastiveCrop \cite{Peng-ContrastiveCrop-2022-CVPR} is a plug-and-play crop module for contrastive cropped pairs, which avoids false positives pairs and similar appearances pairs.

\begin{figure*}[t]
\centering
\includegraphics[width=0.95\textwidth]{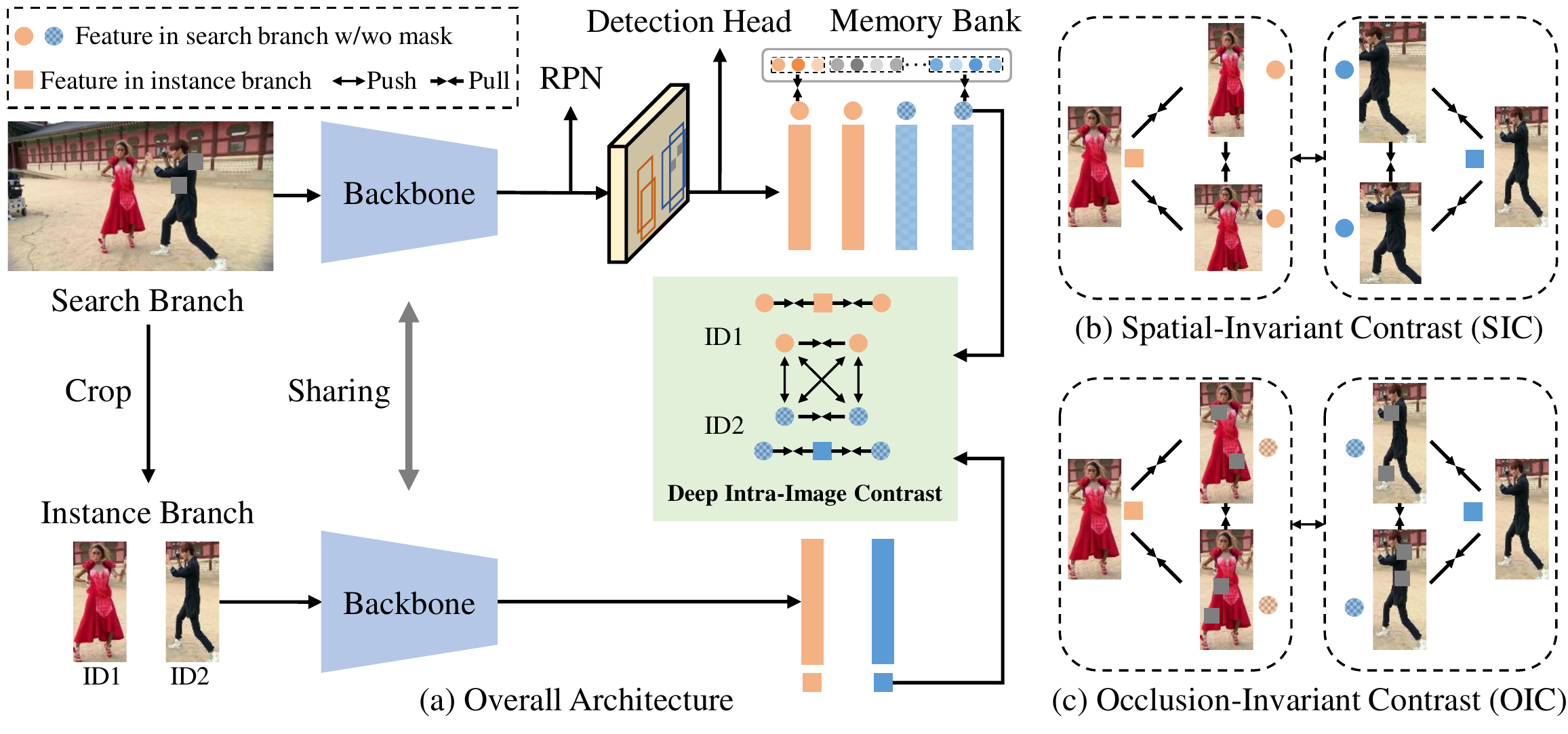}
\caption{Architecture (a) of our deep intra-image contrastive learning (DICL) based method using a Siamese network, which has a search branch and an instance branch. DICL conducts intra-image contrast using two novel modules: spatial-invariant contrast (SIC) and occlusion-invariant contrast (OIC). SIC (b) performs many-to-one contrast in two branches of Siamese network and dense contrasts in all predictions of an image. OIC (c) enhances feature consistency using the masking strategy. }
\label{fig_overall}
\end{figure*}

\section{Our Method}
In this section, we first introduce the motivation, second present the overall architecture of proposed method, third describe two novel modules, and finally give the processes during training and inference.

\subsection{Motivation}\label{sec:motivation}
Contrastive learning has been initially used for weakly supervised one-step person search \cite{Yan-CGPS-2022-AAAI,Han-RSiamNets-2021-ICCV}, including memory-based contrast and intra-image contrast. However, the performance still lags far behind  fully supervised one-step person search.  We argue that one of the key reasons is that these   methods do not fully mine intra-image information to deal with spatial-level and occlusion-level variance of person instance. For example, the detected person under different scenes may have different scales or inaccurate bounding-box localization, which is termed as spatial-level variance. The occlusion-level variance is that the detected person is occluded by other objects sometimes. As a result, it becomes challenging to accurately match the re-id features of same person across various images due to these spatial-level and occlusion-level variance.

To bridge the performance gap between weakly supervised one-step methods and fully supervised one-step methods, we present a novel deep intra-image contrastive learning based method (DICL) to fully exploit instance-level information of each person image considering spatial-level and occlusion-level variance of person instance.

\subsection{Overall Architecture}\label{sec:arch}
Fig. \ref{fig_overall} gives the overall architecture of our  DICL. We adopt the Siamese network that contains a search branch and an instance branch. The search branch  extracts the features of entire image for detection and re-id, while the instance branch extracts the re-id features of fixed-size ground-truth pedestrians cropped and resized from the entire image. The feature extraction is performed based on Faster R-CNN \cite{Ren-FasterRCNN-2017-PAMI}. Different to original Faster R-CNN, we make some modifications as follows. We use the fused feature map of last two layers of backbone for region proposal network (RPN) and RoI head network. The RoI head network consists of one $3\times3$ deformable layer \cite{Dai_DCN_ICCV_2017}, two $3\times3$ convolutional layers, and three parallelled fully-connected layers. In search branch, we first use an RoIAlign layer to extract the features of proposals and second employ RoI head network to perform  classification, regression, and re-id feature generation. In instance branch, we remove the RoI layer and directly extract the re-id features using RoI head network.

Based on the re-id features generated in search branch and instance branch, we conduct intra-image contrastive learning and memory-based contrastive learning during training. For intra-image contrastive learning, we present a novel deep intra-image contrastive learning strategy that includes a spatial-invariant contrast module and an occlusion-invariant contrast module, described below.
For memory-based contrastive learning, we build a cluster-level re-id memory bank using SPCL \cite{Ge-SPCL-2020-NeurIPS}, and supervise the re-id feature learning like  \cite{Yan-CGPS-2022-AAAI}.
The memory bank is initialized  with the averaged re-id features of two branches.

\subsection{Spatial-Invariant Contrast}
\label{spatial-level}

The same person under different cameras appears with the variance in scale. In addition, the detection  cannot guarantee an accurate bounding-box localization like the ground-truth annotation. Therefore, the detected person in different images may have different scales or inaccurate bouding-box localization. We call it as spatial-level variance. The spatial-level variance will affect the accurate re-id matching across images. To deal with this spatial-level variance, we propose a spatial-invariant contrast (SIC) module that performs many-to-one Siamese contrasts between two branches of Siamese network and dense prediction contrasts in search branch. 

The many-to-one Siamese contrast aims to enforce feature consistency between multiple predicted re-id features of a person in search branch and one re-id feature of a ground-truth in instance branch. We assign the predicted re-id features to the person when the corresponding predicted bounding-boxes have the IoU overlap higher than a given threshold with the person ground-truth bounding-box. Assuming that an image has $n^p$ predictions and $n^g$ ground-truths, the re-id features in search branch are represented as $f_i, i=1,..,n^p$, and the re-id features in instance branch are represented as $f_j^g, j=1,...,n^g$. We enforce the  re-id features of the same person in search branch to be consistent with re-id feature of corresponding ground-truth, where the loss of a prediction 
is written as 
\begin{equation}
L_{mto}(i) = 1 - s(f_i, f^{g}_{j(i)}), i=1,...,n^p,
\label{eq:sic} 
\end{equation}
where $s$ represents the cosine distance similarity, and $j(i)$ represents the $i$-th predicted re-id feature corresponds to the $j$-th ground-truth. The dense prediction contrast aims  to enforce the re-id features of the same person be closer and push the re-id features of different persons apart. The loss of dense contrast is formulated by triplet loss with the hard mining strategy \cite{Hermans-hardtri-2017-ArXiv}, which is written as  
\begin{equation}
    \label{onepair}
     L_{tri}(i)=m+\mathop{\max}_{f\in f_i^{+}}d(f_i,f)-\mathop{\min}_{f\in f_i^{-}}d(f_i,f),
\end{equation}
where $d$ is a distance metric function, $f_i^+$ indicates a positive sample set that has the same identity with predicted $f_i$, $f_i^-$ indicates a negative sample set that has  different identities with predicted $f_i$, and $m$ is a distance margin parameter. For a fair comparison, many-to-one Siamese contrast adopts the similar losses used in CGPS and R-SiamNet, and do not explore how to design a better contrastive loss, like NCE.

The many-to-one Siamese contrast is able to enforce the pedestrian re-id features of different scales and bounding-box locations to be consistent with each other, while dense prediction contrast makes the re-id features of the same person tighter with each other meanwhile keeps apart the features with different ids. Therefore, our SIC tends to learn discriminative  scale-invariant and location-invariant, which can provide an accurate re-id matching even when the detected bounding-box is inaccurate or has a large variance in scale.


\subsection{Occlusion-Invariant Contrast}
\label{context-level}

Occlusion is very common in our daily life, which  leads to a negative effect on person re-id. We call it as occlusion-level variance. To deal with this occlusion-level variance, we propose an occlusion-invariant contrast (OIC) module that performs the contrast between masked re-id feature and unmasked re-id feature. There are three different masking strategies, including masking on search branch, masking on instance branch, and masking on both search and instance branches. We observe that the masking on search branch performs better, as shown in Table \ref{tab:noise}. The reason could be  that instance branch without masking can provide consistent and occlusion-free feature, which can guide search branch to learn occlusion-invariant re-id feature.


Given a person in an image, we divide a person bounding-box into $14\times6$ grids, and randomly erase a portion of grids with pixel mean value.  We perform the masking on each person with the probability of 50\%, and mask at most 2 grids. We enforce the  re-id feature of masked person in search branch to be consistent with unmasked re-id feature in instance search, where the loss of a prediction is written as
\begin{equation}\label{contrastiveloss_OIC}
    L_{o}(i) = 1 - s(f^{o}_i, f^{g}_{j(i)}),, i=1,...,n^{o},
\end{equation}
where $o$ represents the prediction of masked person, and $n^o$ represents the number of predictions from occluded persons. The OIC module works based on SIC module. When they are performed simultaneously, OIC module also uses triplet loss between masked or unmasked instances. 

\textbf{Comparison with CGPS \cite{Yan-CGPS-2022-AAAI} and R-SiamNet \cite{Han-RSiamNets-2021-ICCV}:} Our proposed IDCL is related to CGPS and R-SiamNet, but has significant differences and contributions. (i) Our main contribution is a deep intra-image contrastive learning (DICL) framework. Our DICL provides two different contrast strategies, including spatial-invariant contrast (SIC) and occlusion-invariant contrast (OIC). Experimental results in Table \ref{tab:performance_comparison_state-of-the-arts} demonstrate that our DICL significantly surpasses CGPS and R-SiamNet. On PRW, the improvements are 19.3\% and 14.3\% in mAP, and the improvements are 12.9\% and 7.5\% in top-1 accuracy. We hope that our simple and effective DICL will serve as a strong baseline and help ease future research in this open-research area. 
(ii) The SIC in our DICL is a deep spatial-invariant contrast strategy. In contrast, CGPS and R-SiamNet are relatively shallow and do not fully exploit instance spatial information. For example, CGPS suffers from uncropped context interference when performing dense contrasts in search branch, while R-SiamNet ignores the impact of inaccurate bounding-box localization with one-to-one GT Siamese contrast. Experimental results in Table \ref{tab:sic} demonstrate that even a simple combination of CGPS and R-SiamNet (\textit{i.e.,} one-to-one GT Siamese contrast + dense contrasts in search branch) is inferior to our SIC (mAP: 31.4\% vs 33.2\%, top-1 accuracy: 77.3\% vs 79.9\%). 
(iii) Our SIC is a unified framework for spatial-invariant contrastive learning, where the contrast strategies in CGPS and R-SiamNet can be seen as a special case. In addition to SIC, we further introduce an occlusion-invariant contrast (OIC).

\begin{algorithm}[t]
\small
\SetAlgoLined
\SetKwInOut{Input}{Input}
\SetKwInOut{Output}{Output}
\label{alg:algorithm}
\Input{Training images; \\
Bounding-box annotations.} 
\Output{Model of search branch.}
\For{each epoch} {
\textbf{a:} Extract features of all training images with bounding-box annotations.\\
\textbf{b:} Update memory bank and cluster. \\
\For{each mini-batch} {
    \textbf{1:} Randomly add mask patches on a portion of person regions of input image in search branch. \\
    \textbf{2:} Generate the re-id features of search branch and instance branch using Siamese network. \\
    \textbf{3:} Compute the contrast re-id loss by Eq. \ref{loss_total} and detection loss. \\
    \textbf{4:} Update the memory bank features. \\
    }
}
\caption{Deep Into-image Contrastive learning Algorithm for weakly supervised person search}
\label{alg:dicl}
\end{algorithm}

\subsection{Training and Inference}
\label{sec_loss}

During training, we build the initialized  re-id memory bank by detecting all the training images and generating the re-id feature of each person with the averaged re-id feature of two branches. At the beginning of training, we perform the cluster on memory bank. At each iteration, the cluster-level memory bank is used for memory-based contrastive learning to supervise the re-id feature learning like \cite{Yan-AnchorFree-2021-CVPR}.
In addition to memory-based contrast, we  perform deep intra-image contrastive learning using the Siamese network. In search branch, we randomly add the mask on a portion of pedestrian region and perform the forward for detection and re-id feature generation. We keep the re-id features of all the $n$ predicted bounding-boxes, where there are $n^p$ re-id features without image masking and $n^o$ re-id features with image masking. In instance branch, we crop and resize the ground-truth image regions into the fixed sizes and perform the forward to generate the re-id features. With the re-id features in both branches, we perform spatial-invariant contrast and occlusion-invariant contrast. The contrast loss of our method can be written as 
\begin{equation}\label{loss_total}
    L_{c} = \frac{1}{n^p}\sum_i L_{mto}^i + \frac{1}{n^{o}}\sum_i L_{o}^i +\frac{1}{n}\sum_i L_{tri}^i+L_{oim},
\end{equation}
where  $L_{oim}$ is the OIM loss for memory-based contrast.
Then, the overall loss can be written as $L_{all} = L_{c}+L_{det}$, where   $L_{det}$ is the detection loss. We summarize our DICL for weakly supervised person search in Algorithm \ref{alg:dicl}.


During inference, we only keep the search branch in Siamese network for detection and re-id feature generation. We first employ search branch to perform detection and re-id feature generation on query image and a set of gallery images, respectively. After that, we set  the re-id feature  having the maximum overlap with the ground-truth bounding-box as re-id feature of each person.  Finally, we calculate re-id feature similarities of query person and persons in gallery images, and select the matched persons according to the similarity scores.

\section{Experiments}

\subsection{Datasets and Settings}

\textbf{CUHK-SYSU} \cite{Xiao-cuhk-2017-CVPR} is a large-scale person search dataset, including street  and movie snapshots. It contains 18,184 images, 96,143 annotated pedestrian bounding-boxes, and  8,432 person identities. The training set contains 11,206 images, 55,272 pedestrian bounding-boxes,  and 5,532 person identities, while the test set contains 6,978 images,  40,871 pedestrian bounding-boxes, and 2,900 person identities. During inference, the different queries have different gallery images, and the size of gallery images ranges from 50 to 4,000. If no special illustration, the gallery size is set to 100 as the default like most person search methods.

\textbf{PRW} \cite{Zheng-prw-2017-CVPR} is captured in Tsinghua university using six cameras at different positions. Five cameras have a resolution of $1,080\times1,920$ pixels and another one is $576\times720$ pixels. There are 11,816 images, 43,110 pedestrian bounding-boxes, and 932 person identities. The training set contains 5,704 images, 18,048 pedestrian bounding-boxes and 482 person identities. The test set has 6,112 images, 25,062  pedestrian bounding-boxes, and 450 person identities. 

\textbf{Evaluation Protocol.} We employ two standard evaluation metrics widely adopted in person search, including mean averaged precision (mAP) and top-1 accuracy.

\subsection{Implementation Details}
\label{Implementation Details}

We implement our  method on  open-source library mmdetection \cite{Chen-MMDetection-2019-ArXiv}. We adopt ResNet50 pre-trained on the large-scale ImageNet-1K dataset \cite{Deng-Imagenet-2009-CVPR}, and use the fused feature map of last two layers for RPN and RoI head network. The fused feature map has the resolution of $H/16\times W/16\times1024$ pixels. Search branch employs RoIAlign to generate the feature map of $14\times6\times1024$ pixels. Instance branch crops and resizes the pedestrian image into the fixed size of $224\times96$ pixels, so that the fused feature map in instance branch has the resolution of $14\times6\times1024$ pixels. The final fully-connected layer generates  256-$d$ vectors for  re-id feature. We adopt  the cluster hyper-parameters used in SPCL \cite{Ge-SPCL-2020-NeurIPS} to generate cluster-level re-id memory bank. In addition, we employ the prior that the persons in an image belong to different identities during cluster like  \cite{Jia-CGUA-2022-ArXiv,Yan-CGPS-2022-AAAI}.


Our experiment is conducted on a single NVIDIA GeForce RTX 3090 GPU with SGD optimizer. The mini-batch is set as four images and the image scale is fixed to $1,500\times900$ pixels. There are 26 epochs in total, where the initial learning rate is set as 0.001 and decreased by a factor of 0.1 after epoch 16 and 22. During inference, the image is resized to $1,500\times900$ pixels.

\begin{table}[t]
\caption{Impact of different modules in our DICL, including spatial-invariant contrast (SIC) and occlusion-invariant contrast (OIC), for weakly supervised person search on PRW test set.}
\begin{center}
\setlength{\tabcolsep}{4.0mm}{
\begin{tabular}{cc|c|c}
\hline
\multicolumn{2}{c|}{Method (Weakly supervised)}&\multicolumn{2}{c}{re-id}\cr
\hline
SIC & OIC & mAP&top-1\cr
\hline
&&25.6&73.2\cr
\checkmark&&33.2&79.9\cr
\checkmark&\checkmark&{\bf 35.5}&{\bf 80.9}\cr
\hline
\end{tabular}}
\end{center}
\label{tab:overall}
\end{table}

\begin{table}[t]
\caption{Impact of different modules in our DICL, including spatial-invariant contrast (SIC) and occlusion-invariant contrast (OIC), for fully supervised person search on PRW test set.}
\begin{center}
\setlength{\tabcolsep}{4.2mm}{
\begin{tabular}{cc|c|c}
\hline
\multicolumn{2}{c|}{Method (Fully supervised)}&\multicolumn{2}{c}{re-id}\cr
\hline
SIC & OIC&mAP&top-1\cr
\hline
&&43.6&78.8\cr
\checkmark&&48.8&82.2\cr
\checkmark&\checkmark&{\bf 49.6}&{\bf 82.4}\cr
\hline
\end{tabular}}
\end{center}
\label{tab:overall_fully}
\end{table}

\begin{table}[t]
\caption{Impact of integrating many-to-one Siamese contrast and dense contrast in our spatial-invariant contrast (SIC) module.}
\begin{center}
\setlength{\tabcolsep}{3.5mm}{
\begin{tabular}{c|cc|cc}
\hline
Num. & Many-to-one & Dense &mAP&top-1\cr
\hline
(a) &  & &25.6&73.2\cr
(b) & \checkmark & &32.2 &79.6 \cr
(c) & & \checkmark &  31.4&77.3 \cr
(d) & \checkmark &\checkmark& \bf{33.2}&\bf{79.9}\cr
\hline
\end{tabular}}
\end{center}
\label{tab:sic}
\end{table}

\begin{table}[t]
\caption{Impact of different Siamese contrast settings in our spatial-invariant contrast (SIC) module. Ground-truth indicates ground-truth bounding-box of a person, while positives indicate all the predictions belongs to a person.}
\begin{center}
\setlength{\tabcolsep}{2.0mm}{
\begin{tabular}{c|c|c|cc}
\hline
Num. & Search branch&Instance branch&mAP&top-1\cr
\hline
(a) &Ground-truth &Ground-truth&25.6&73.2\cr
(b) &Positives &Positives&31.2&79.7\cr
(c) &Positives&Ground-truth&\bf{33.2}&\bf{79.9}\cr
\hline
\end{tabular}}
\end{center}
\label{tab:translation}
\end{table}

\begin{table}[t]
\caption{Impact of masking strategy on different branches  in our occlusion-invariant contrast (OIC) module.}
\begin{center}
\setlength{\tabcolsep}{4.5mm}{
\begin{tabular}{c|c|cc}
\hline
Num. & Masking strategy &mAP&top-1\cr
\hline
(a) & None &33.2&79.9\cr
(b) &  Search branch&\bf{35.5}&\bf{80.9}\cr
(c) &  Instance branch&33.1&79.5\cr
(d) &  Two branches&32.7&80.1\cr
\hline
\end{tabular}}
\end{center}
\label{tab:noise}
\end{table}

\begin{table}[t]
\caption{Impact of masking grids  in our occlusion-invariant contrast (OIC) module.}
\begin{center}
\setlength{\tabcolsep}{4.5mm}{
\begin{tabular}{c|cccc}
\hline
Num. of grids & 0 & 2 & 4 & 6\cr
\hline
mAP & 33.2 &35.5&35.4 & 33.0\cr
Top-1 accuray &  79.9 & 80.9 & 80.3 & 80.0\cr
\hline
\end{tabular}}
\end{center}
\label{tab:ratio}
\end{table}


\begin{table}[t]
\caption{Comparison of w/wo occlusion-invariant contrast (OIC) module under original and masked PRW test set.}
\begin{center}
\setlength{\tabcolsep}{4.0mm}{
\begin{tabular}{c|cc|cc}
\hline
\multirow{2}{*}{Test set} & \multicolumn{2}{c|}{without OIC}&\multicolumn{2}{c}{with OIC}\cr
\cline{2-5}
& mAP &top-1 & mAP &top-1\cr
\hline
Original &33.2 & 79.9&35.5 & 80.9\cr
Masked &15.0 & 60.2&20.9 & 67.6\cr
\hline
$\triangle$ (drop) &18.2 & 19.7&14.6 & 13.3\cr
\hline
\end{tabular}}
\end{center}
\label{tab:noise_test}
\end{table}

\subsection{Ablation Study} 
Here, we perform ablation study on PRW dataset to demonstrate the effectiveness of proposed method DICL.

\textbf{Impact of different modules in DICL} Table \ref{tab:overall} shows the impact of our deep intra-image contrastive learning (DICL) for weakly supervised person search on PRW test set. The DICL contains a spatial-invariant contrast (SIC) module and an occlusion-invariant contrast (OIC) module. The baseline (a) only employs the ground-truth bound-box for intra-image contrast learning like \cite{Han-RSiamNets-2021-ICCV}.  When integrating our SIC module into the baseline, it provides 7.6\% improvement on mAP and 6.7\% improvement on top-1 accuracy. When  integrating our OIC module into them, it further provides 2.3\% improvement on mAP and 1.0\% improvement on top-1 accuracy. Overall, our DICL provides 9.9\% improvement on mAP and 7.7\% improvement on top-1 accuracy. It demonstrates that our deep intra-image contrast learning is effective on weakly supervised person search.

In addition, we show the impact of our DICL  for fully supervised person search in Table \ref{tab:overall_fully}. We build the re-id memory bank based on identity annotations, instead of unsupervised cluster. Our DICL achieves the mAP of 49.6\% and the top-1 accuracy of 82.4\%, which outperforms the baseline by 6.0\% and 3.6\%.  The related BINet \cite{Dong-BINet-2020-CVPR} also adopts the Siamese framework for fully supervised person search, which achieves the mAP of 45.3\% and the top-1 accuracy of 81.7\%. Therefore, our DICL outperforms BINet by 4.3\% on mAP and 0.7\% on top-1 accuracy. It demonstrates the effectiveness of our deep intra-image contrast learning on fully supervised person search. Because this paper mainly focuses on weakly supervised settings, we do not add fully-supervised results in Table \ref{tab:performance_comparison_state-of-the-arts}.
Meanwhile, our method has a larger improvement on weakly supervised person search, which demonstrates that intra-image contrast is more important for weakly supervised person search.

\textbf{Impact of different contrasts in SIC} We show the impact of integrating many-to-one Siamese contrast and dense prediction contrast in Table \ref{tab:sic}. Many-to-one Siamese contrast denotes the contrasts between multiple predictions of a person in search branch and one ground-truth prediction of a person in instance branch. Dense prediction contrast denotes that the contrasts between all the predictions in search branch. When integrating single many-to-one Siamese contrast between two branches (b) or dense prediction contrast in search branch (c), it can significantly improve performance, respectively. When integrating these two contrasts together (d), it  achieves the best performance. In addition, the method (c) can be treated as the combination of CGPS and R-SiamNet, which is inferior to our method (d). Therefore, our SIC methods outperform the simple combination of two weakly supervised methods CGPS and R-SiamNet.

Table \ref{tab:translation} shows the impact of  different Siamese contrast settings. When we perform one-to-one Siamese contrast of ground-truth like \cite{Han-RSiamNets-2021-ICCV}, it (a) achieves 25.6\% on mAP and 73.2\% on top-1 accuracy.  Further, we perform one-to-one Siamese contrasts of all positives, it (b) achieves 31.2\% on mAP and 79.7\% on top-1 accuracy. That provides a substantial improvement on the baseline with  more diverse training samples. When we perform many-to-one Siamese contrasts between  all positive samples and their corresponding ground-truth, it (c) achieves best performance. Compared to the strategy (b), the strategy (c) is more suitable to learn scale-invariant and localization-invariant re-id feature.

\begin{table}[t]
\caption{Comparison with state-of-the-art methods, including fully supervised and weakly supervised approaches on CUHK-SYSU and PRW test sets.}
\label{tab:performance_comparison_state-of-the-arts}
\centering
\resizebox{1\columnwidth}{!}{
\begin{tabular}{c|c|cc|cc}
\hline
\multicolumn{2}{c}{\multirow{2}*{Methods}}&
\multicolumn{2}{|c|}{CUHK-SYSU}&\multicolumn{2}{c}{PRW}\cr
\cline{3-6} 
\multicolumn{2}{c|}{}&mAP&top-1&mAP&top-1\cr
\hline
\multirow{9}* {\rotatebox{90}{Two-step}}
&\multicolumn{5}{l}{\textit{Fully supervised: }}\cr
\cline{2-6} 
&IDE \cite{Zheng-prw-2017-CVPR}&-&-&20.5&48.3\cr
&MGTS \cite{Chen-MGTS-2018-ECCV}&83.0&83.7&32.6&72.1\cr
&CLSA \cite{Lan-CLSA-2018-ECCV}&87.2&88.5&38.7&65.0\cr
&RDLR \cite{Han-RDLR-2019-ICCV}&93.0&94.2&42.9&70.2\cr
&IGPN \cite{Dong-IGPN-2020-CVPR}&90.3&91.4&47.2&87.0\cr
&TCTS \cite{Wang-TCTS-2020-CVPR}&93.9&95.1&46.8&87.5\cr
\cline{2-6} 
&\multicolumn{5}{l}{\textit{Weakly supervised: }}\cr
\cline{2-6} 
&CGUA \cite{Jia-CGUA-2022-ArXiv}&91.0&92.2&42.7&86.9\cr
\hline
\hline
\multirow{22}*{\rotatebox{90}{One-step}}
&\multicolumn{5}{l}{\textit{Fully supervised: }}\cr
\cline{2-6} 
&OIM \cite{Xiao-cuhk-2017-CVPR}&75.5&78.7&21.3&49.4\\
&IAN \cite{Xiao-IAN-2019-PR}&76.3&80.1&23.0&61.9\cr
&NPSM \cite{Liu-NPSM-2017-ICCV}&77.9&81.2&24.2&53.1\cr
&RCAA \cite{Chang-RCAA-2018-ECCV}&79.3&81.3&-&-\cr
&CTXG \cite{Yan-CTXG-2019-CVPR}&84.1&86.5&33.4&73.6\cr
&QEEPS \cite{Bharti-QEEPS-2019-CVPR}&88.9&89.1&37.1&76.7\cr
&HOIM \cite{Chen-HOIM-2020-AAAI}&89.7&90.8&39.8&80.4\cr
&BINet \cite{Dong-BINet-2020-CVPR}&90.0&90.7&45.3&81.7\cr
&NAE \cite{Chen-NormAware-2020-CVPR}&91.5&92.4&43.3&80.9\cr
&PGA \cite{Kim-Prototype-Guided-2021-CVPR}&90.2&91.8&42.5&83.5\cr
&AlignPS \cite{Yan-AnchorFree-2021-CVPR}&93.1&93.4&45.9&81.9\cr
&DMRNet \cite{Han-DMRNet-2021-AAAI}&93.2&94.2&46.9&83.3\cr
&AGWF \cite{Han-AGWF-2021-ICCV}&93.3&94.2&53.3&87.7\cr
&SeqNet \cite{Li-seqnet-2021-AAAI}&93.8&94.6&46.7&83.4\cr
&BUFF \cite{Yang-CSVT-2022-PR}&91.6&92.2&44.9&86.3\cr
&OIMNet++\cite{Lee-OIMNet++-2022-ECCV}&93.1&94.1&47.7&84.8\cr
&PSTR \cite{Cao-PSTR-2022-CVPR}&93.5&95.0&49.5&87.8\cr
&COAT \cite{Yu-COAT-2022-CVPR}&94.2&94.7&53.3&87.4\cr
\cline{2-6} 
&\multicolumn{5}{l}{\textit{Weakly supervised: }}\cr
\cline{2-6} 
&CGPS \cite{Yan-CGPS-2022-AAAI}&80.0&82.3&16.2&68.0\cr
&R-SiamNet \cite{Han-RSiamNets-2021-ICCV}&86.0&87.1&21.2&73.4\cr
&{\bf DICL (Ours)}&{\bf87.4}&{\bf88.8}&{\bf35.5}&{\bf 80.9}\cr
\hline
\end{tabular}}
\end{table}

\textbf{Impact of  masking strategy in OIC}  Table \ref{tab:noise} shows the impact of the masking strategy on different branches of Siamese network. It achieves the best performance by performing the masking only on search branch. The masking strategies on instance branch or on both two branches do not improve the performance on mAP. We argue the reason is that instance branch aims to guide re-id feature to be consistent, while the masking strategy on instance branch will make it difficult to learn the consistent features. We also show the impact of masking grids in Table \ref{tab:ratio}. We observe that masking 2 grids has best performance.

To show the ability of our OIC to deal with occlusion variance, we generate a masked test set by randomly masking 20\% pixels of test images. Table \ref{tab:noise_test} compares the performance on original test test or masked test set. When using masked test set, all the methods have the performance drop. We observe that it has less performance drop when using our OIC module. It demonstrates that our occlusion-invariant contrast can better deal with occlusion variance.

\subsection{Comparison With State-Of-The-Art Methods}

We compare our proposed method DICL with the state-of-the-art methods, including fully supervised  and weakly supervised approaches. They can be divided into two-step approaches and one-step approaches.

\textbf{Comparison on CUHK-SYSU} In Table \ref{tab:performance_comparison_state-of-the-arts}, we compare our  DICL with some  state-of-the-art approaches with the gallery size of 100, including fully-supervised and weakly-supervised methods. Our method DICL achieves the mAP of 87.4\% and top-1 accuracy of 88.8\%, which is superior to current weakly supervised one-step  person search methods. CGPS \cite{Yan-CGPS-2022-AAAI} has the mAP of 80.0\% and top-1 accuracy of 82.3\%, while R-SiamNet \cite{Han-RSiamNets-2021-ICCV} has the mAP of 86.0\% and top-1 accuracy of 87.1\%. Our proposed method outperforms CGPS by 7.4\% on mAP and 6.5\% on top-1 accuracy, and outperforms R-SiamNet by 1.4\% on mAP and 1.7\% on top-1 accuracy. In addition, our method outperforms some fully supervised one-step methods such as RCAA \cite{Chang-RCAA-2018-ECCV} and CTXG \cite{Yan-CTXG-2019-CVPR}. The weakly-supervised two-step CGUA \cite{Jia-CGUA-2022-ArXiv} has the mAP of 91.0\% and top-1 accuracy of 92.2\%, 
 which uses two independent networks for detection and re-id, and focuses on memory-based contrast.  Compared to weakly-supervised two-step CGUA, our one-step DICL only utilizes a single network for detection and re-id, and focuses on deep intra-image contrast. Therefore, our DICL is a simple framework for end-to-end person search. 

\begin{figure}[t]
\centering
\includegraphics[width=0.4\textwidth]{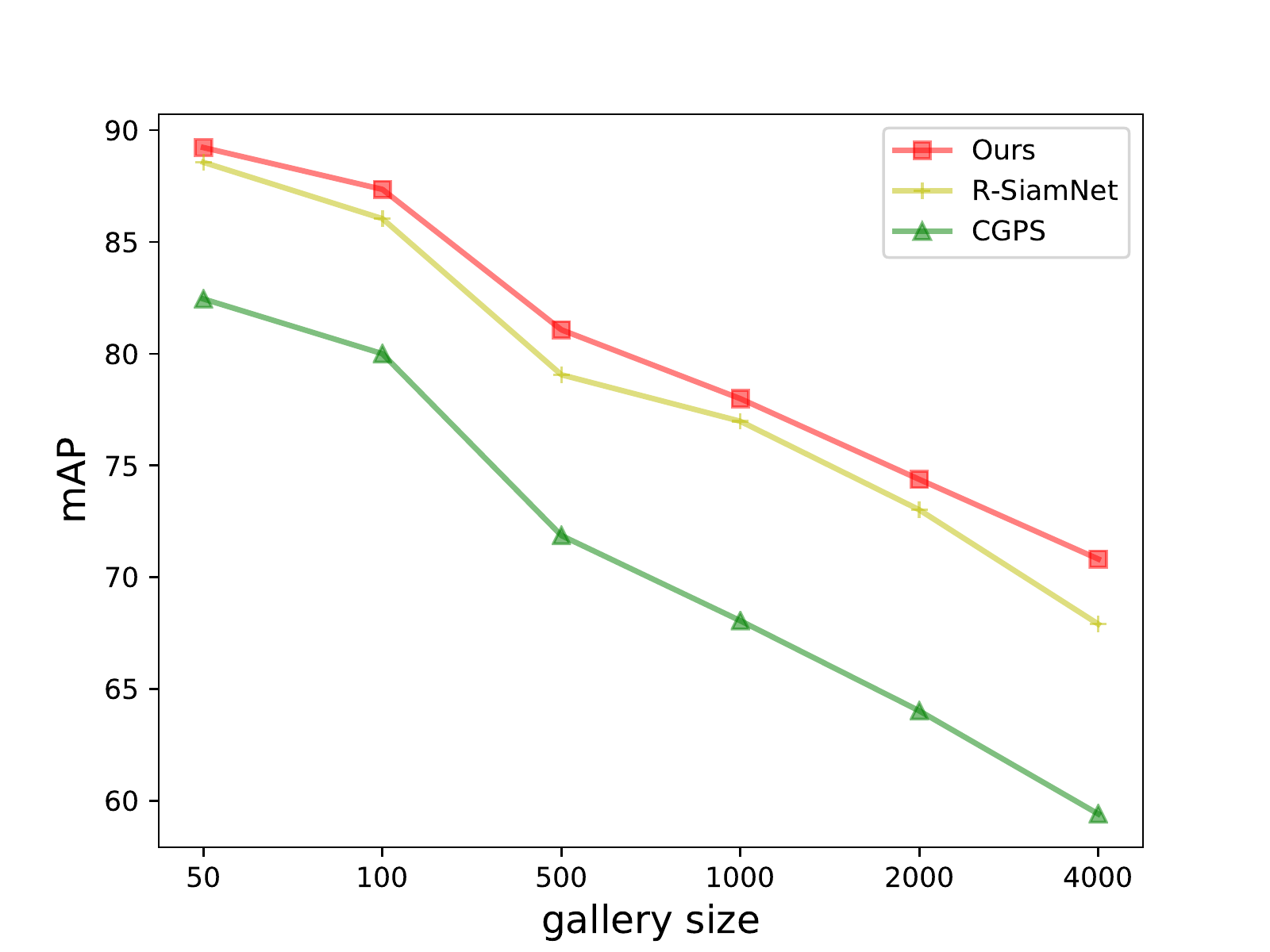}
\caption{Comparison with different weakly supervised one-step methods under different gallery sizes on CUHK-SYSU test set.}
\label{fig_gallery}
\end{figure}

Fig \ref{fig_gallery} further  presents the performance under different gallery sizes of [50, 100, 500, 1000, 2000, 4000]. The result shows that our method works  better than weakly supervised one-step methods, especially under larger gallery sizes, which indicates that our method is  capable of searching targets in large and challenging scenes.

\textbf{Comparison on PRW} We also compare our method with  state-of-the-art approaches on PRW test set in Table \ref{tab:performance_comparison_state-of-the-arts}. Compared to CUHK-SYSU, PRW test set has a larger gallery size (6,112) and is  challenging. Our method achieves a mAP of 35.5\% and a top-1 accuracy of 80.9\%, which outperforms  CGPS by 19.3\% on mAP and 12.9\% on top-1 accuracy, and outperforms  R-SiamNet by 14.3\% on mAP and 7.5\% on top-1 accuracy. In addition, our method  outperforms  fully supervised methods MGTS \cite{Chang-RCAA-2018-ECCV}  and CTXG \cite{Yan-CTXG-2019-CVPR} by 2.9\% and 2.1\% on mAP respectively.

\begin{figure*}[t]
\centering
\includegraphics[width=0.98\textwidth]{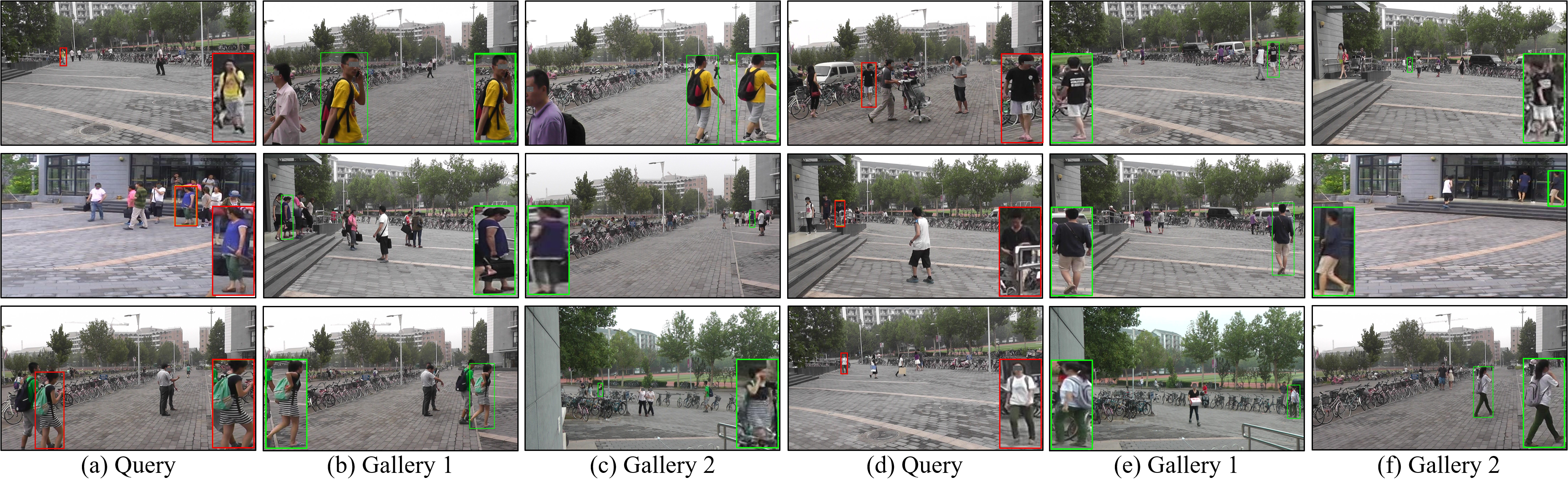}
\caption{Qualitative results of our method on PRW test set, where the red box represents the query and the green box represent search result in the gallery image. Our method finds the query persons with various views and scales.}
\label{fig_visprw}
\end{figure*}

\begin{figure*}[t]
\centering
\includegraphics[width=0.98\textwidth]{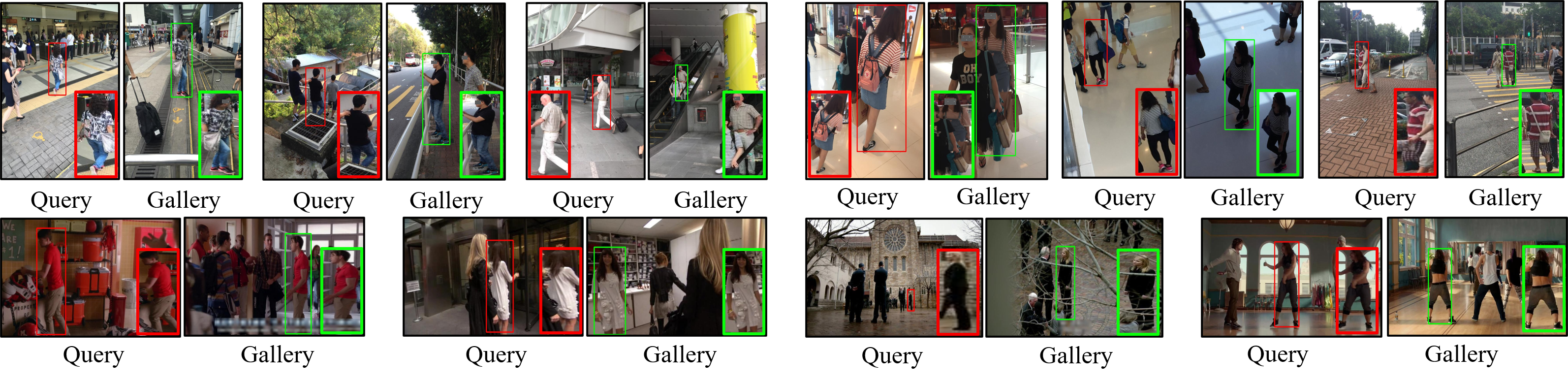}
\caption{Qualitative results of our method on CUHK-SYSU test set, where the red box represents the query and the green box represent
search result in the gallery image. Our method matches the query persons in different scenes.}
\label{fig_viscuhk}
\end{figure*}

\begin{figure}[t]
\centering
\includegraphics[width=0.48\textwidth]{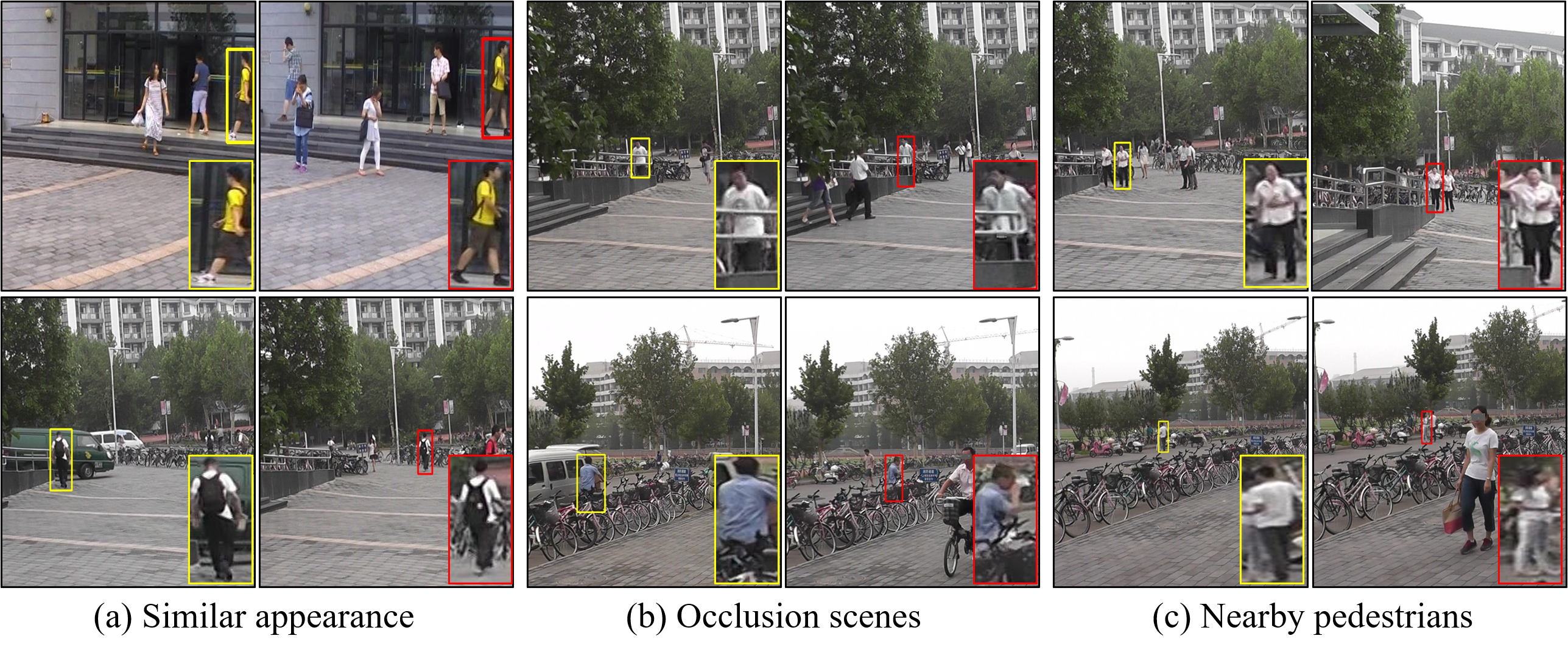}
\caption{Some failure cases of our method. We observe that it is still challenging to detect and recognize the persons at heavy occlusion, extreme low-light conditions, etc.}
\label{fig_failure}
\end{figure}

\textbf{Qualitative results} Fig. \ref{fig_visprw} and Fig. \ref{fig_viscuhk}  give some qualitative results of our DICL on PRW and CUHK-SYSU test sets. Our method successfully detects and recognizes the query persons in different gallery images, where the persons have different poses, scales, background and visibility. For example, as shown in the last row of Fig. \ref{fig_visprw}, the girl changes greatly in position, scale and visible part.  We also observe that our method occasionally struggles at similar appearance, heavy occlusion, and nearby pedestrian interference in Fig. \ref{fig_failure}. Therefore, it is necessary to address these issues in future.

\section{Conclusion}

In this paper, we have proposed a novel deep intra-image contrastive learning (DICL) framework for weakly supervised one-step  person search. We  employed a Siamese network with a search branch and an instance branch to learn discriminative features to deal with spatial-level and occlusion-level variances. Specifically, we  introduced two novel intra-image contrast modules: spatial-invariant contrast (SIC) and occlusion-invariant contrast (OIC). SIC performs many-to-one Siamese contrasts between two branches of Siamese network and dense contrasts in all the predictions in search branch, which can learn discriminative scale-invariant and location-invariant re-id features. OIC preforms the masking strategy on search branch and improves feature consistency to occlusion variance. Extensive experiments on two datasets demonstrate the effectiveness of our method. 

Currently, our method mainly focuses on deep intra-image contrast, which does not consider to improve memory-based contrast. In future, we will exploit how to improve the unsupervised cluster for memory-based contrast.

\end{document}